%% file: main.tex
\documentclass[10pt,twocolumn,letterpaper]{article}

\usepackage{cvpr}   
\input{preamble}
\definecolor{cvprblue}{rgb}{0.21,0.49,0.74}
\usepackage[pagebackref,breaklinks,colorlinks,allcolors=cvprblue]{hyperref}
\usepackage{geometry}
\usepackage{amsmath}
\usepackage{amsfonts}
\usepackage{bm}
\usepackage{graphicx}
\usepackage{booktabs}
\usepackage{algorithm}
\usepackage{algpseudocode}
\usepackage{enumitem}
\usepackage{multirow}
\usepackage{caption}
\usepackage{hyperref}
\usepackage[table]{xcolor} 
\def\paperID{37212} 
\def\confName{CVPR}
\def\confYear{2026}

\newcommand{\diag}{\mathrm{diag}}

\title{Probabilistic Concept Graph Reasoning for Multimodal Misinformation Detection}

\author{Ruichao Yang$^{1}$, Wei Gao$^{2}$, Xiaobin Zhu$^{1}$\thanks{Corresponding authors.} , Jing Ma$^{3*}$, Hongzhan Lin$^{3}$, \\
Ziyang Luo$^{3}$, Bo-Wen Zhang$^{1*}$, Xu-Cheng Yin$^{1}$\\
$^{1}$ University of Science and Technology Beijing, \\$^{2}$ Singapore Management University,
$^{3}$ Hong Kong Baptist University\\
\texttt{\{yangruichao,zhuxiaobin\}@ustb.edu.cn}, 
\texttt{weigao@smu.edu.sg} \\
\texttt{\{majing,cshzlin,cszyluo\}@comp.hkbu.edu.hk}
}

\begin{document}
\maketitle
\input{sec/0_abstract}    
\input{sec/1_intro}

\input{sec/2_relatedwork}
\input{sec/3_problemform}
\input{sec/4_method}
\input{sec/5_experiment}
\input{sec/6_conclusion}

\section*{Acknowledgments}
This research is supported by the National Science Fund for Distinguished Young Scholars (No. 62125601), National Natural Science Foundation of China (No. 62576031), National Natural Science Foundation of China Young Scientists Fund (No. 62206233), and National Research Foundation, Singapore under its AI Singapore Programme (AISG Award No: AISG3-RP-2024-035).

{
    \small
    \bibliographystyle{ieeenat_fullname}
    \bibliography{main}
}
\input{sec/X_suppl}

\end{document}

%% file: sec/0_abstract.tex
\begin{abstract}
Multimodal misinformation poses an escalating challenge that often evades traditional detectors, which are opaque black boxes and fragile against new manipulation tactics. We present Probabilistic Concept Graph Reasoning (PCGR), an interpretable
and evolvable framework that reframes multimodal misinformation detection (MMD) as structured and concept-based reasoning. PCGR follows a build-then-infer paradigm, which first constructs a graph of human-understandable concept nodes, including novel high-level concepts automatically discovered and validated by multimodal large language models (MLLMs), and then applies hierarchical attention over this concept graph to infer claim veracity. This design produces interpretable reasoning chains linking evidence to conclusions. Experiments demonstrate that PCGR
achieves state-of-the-art MMD accuracy and robustness to emerging manipulation types, outperforming prior methods in both coarse detection and fine-grained manipulation recognition. 
\end{abstract}
\vspace{-0.2cm}

%% file: sec/1_intro.tex
\section{Introduction}
\label{sec:intro}
Online platforms increasingly blend text~\cite{ yang2023wsdms,lin2025fact, lin2026towards} with visuals (e.g., photos, memes, videos)~\cite{lin2024goat, chen2025memearena, lin2025explainhm++} to amplify persuasion through cross-modal cues, creating fertile ground for multimodal misinformation~\cite{wang2024mfc}. A recent example illustrates this threat: a fabricated story, paired with a striking image, falsely claimed that ``aliens would attack Earth in November" and went viral on social media.
Because ``seeing is believing'', many readers, especially non-experts, struggle to discern misinformation in such multimodal formats. This calls for interpretable, automated systems capable of multimodal misinformation detection (MMD) to preserve a trustworthy information ecosystem.

Early research primarily relied on text~\cite{ma2018rumor,shu2019defend,fu2021sdg}, overlooking the visual signals that often reinforce false claims. Later work showed that compelling images can increase the perceived credibility of misleading text~\cite{cao2020exploring}, motivating multimodal fusion methods that integrate image and text features via early/late fusion or co-attention~\cite{wu2021multimodal,jing2023multimodal,zhou2023multimodal,singhal2022leveraging}. However, such end-to-end models are opaque, providing little insight into how decisions are made. Recent interpretable variants, such as multi-view semantic fusion~\cite{zeng2023multi} and knowledge-adaptive multi-expert frameworks~\cite{shen2025gamed}, improve transparency, yet still rely on post-hoc or discrete rationales rather than mechanistic reasoning.

\begin{figure}
    \centering
    \includegraphics[width=0.99\linewidth]{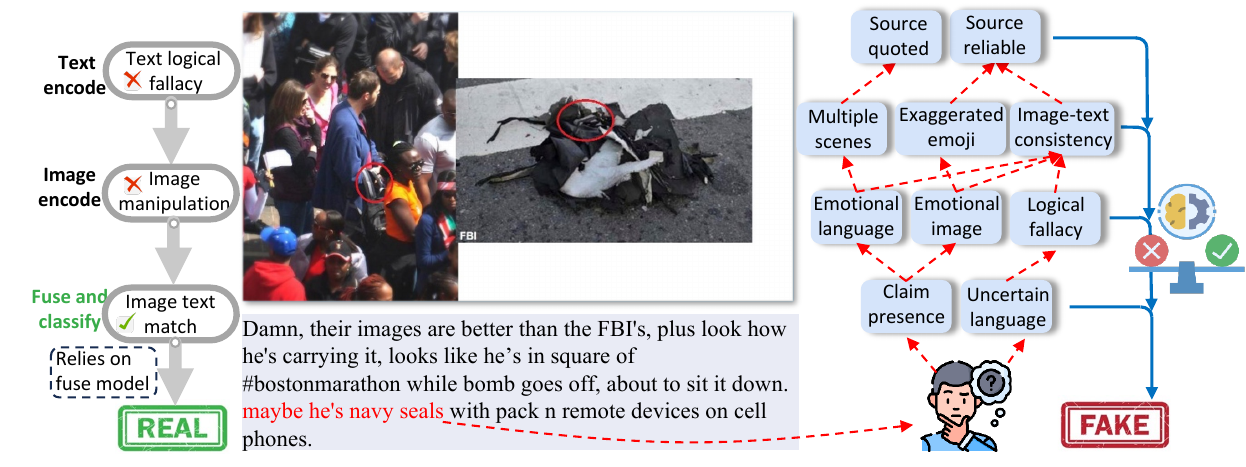}
    \vspace{-0.6cm}
\caption{Human vs. model reasoning on a misinformation example. \emph{Left}: Conventional multimodal detectors fuse image-text features and may misclassify superficially aligned content (e.g., a fake bombing claim supported by a consistent but unrelated photo). \emph{Right}: A concept-based reasoning process, used by human fact-checkers and our model, decomposes reasoning into concept-level checks. Each concept yields a soft judgment, collectively leading to the final verdict.}
\vspace{-0.5cm}
    \label{fig:MotivFig}
\end{figure}

In contrast, human fact-checkers approach multimodal claims through structured reasoning. As shown in Figure~\ref{fig:MotivFig}, a fact-checker decomposes a claim into a series of concept-level evaluations: (i) identifying the core claim and hedging cues (e.g., “\emph{maybe} he’s navy seals”); (ii) examining image authenticity and semantic consistency; (iii) verifying alignment between visual and textual context; and (iv) assessing source credibility and factual details. Each intermediate \emph{concept} contributes to a final veracity decision. This reasoning is both \textit{explainable} and \textit{adaptive}, allowing humans to integrate new inconsistencies as they emerge.

Motivated by this observation, we reformulate MMD as structured reasoning over explicit \textit{concepts}, rather than direct classification. Recent works have explored similar directions by integrating knowledge bases or expert modules for explainability~\cite{wang2020fake,shen2025gamed}, or generating natural-language rationales via MLLMs~\cite{Tahmasebi2024LVLM}. Yet, these either depend on fixed concept sets that hinder generalization or produce post-hoc explanations detached from inference. We instead propose a framework where reasoning is an integral
part of the model architecture, capable of adapting to new manipulation tactics.

Building such a system entails three challenges: (1) expert-defined concepts are domain-specific and hard to generalize; (2) human reasoning is inherently probabilistic, unlike typical hard-decision models; and (3) datasets vary in granularity, from binary veracity to fine-grained manipulation labels.  
To address these issues, we propose Probabilistic Concept Graph Reasoning (PCGR), an interpretable and evolvable framework that models MMD as concept-based probabilistic reasoning. PCGR leverages MLLMs to automatically expand a hierarchical concept set, encodes each multimodal instance into this concept space, and constructs a layered probabilistic graph. Reasoning then proceeds via ``soft'' Graph-of-Thought (GoT) inference~\cite{10.1609/aaai.v38i16.29720}, integrating concept probabilities across layers to yield the final verdict. This produces a transparent and auditable reasoning trail, where concept states (e.g., ``image is out-of-context'', ``text uses hyperbole'', etc.) collectively form the evidence chain supporting the decision.

Our main contributions are summarized as follows:
\begin{itemize}
\item We propose an automatic concept growth procedure that leverages MLLMs to discover and integrate new reasoning concepts, enabling PCGR to adapt to emerging manipulation tactics without costly retraining.
\item We introduce a layered Probabilistic Concept Graph that models MMD as ``soft'' hierarchical reasoning, producing interpretable and verifiable evidence chains.
\item We design a hierarchical attention mechanism that aggregates uncertain concept-level judgments across graph layers, supporting both coarse- and fine-grained manipulation type predictions. \looseness=-1
\item We validate PCGR across multiple benchmark datasets, showing consistent improvements over state-of-the-art MMD models in both accuracy and explainability. Code is released at \url{https://github.com/2302Jerry/pcgr}.
\end{itemize}

  
  


%% file: sec/2_relatedwork.tex
\section{Related Work}
\label{sec:related}

\textbf{Multimodal Misinformation Detection (MMD).}
Research on MMD generally follows two main paradigms: \textit{black-box end-to-end models} and \textit{mechanism-driven models}. End-to-end approaches typically fuse textual and visual representations using pre-trained encoders or transformer architectures~\cite{ijcai2025p891,jin2017multimodal,khattar2019mvae,huangexposing,zhang2024escnet,li2025entity}. While these models achieve strong benchmark performance, their decision processes remain opaque. Recent variants enhance cross-modal alignment and robustness through hierarchical attention~\cite{jin2017multimodal}, contrastive learning~\cite{wang2023ccl}, or synthetic data augmentation~\cite{zeng-etal-2024-multimodal,su2025dynamic}.  
Mechanism-guided models, by contrast, aim to explain predictions via intermediate reasoning components such as manipulation types~\cite{huang-etal-2024-miragenews} or retrieved evidence~\cite{verga-etal-2021-adaptable,pang2025beyond}. Despite improving transparency, they often rely on fixed fine-grained label sets that limit adaptability to novel misinformation tactics. Datasets such as \textsc{MiRAGeNews}~\cite{huang-etal-2024-miragenews} and \textsc{NewsCLIPings}~\cite{luo-etal-2021-newsclippings} enable large-scale multimodal training and evaluation, but domain gaps persist between synthetic and real-world misinformation~\cite{zeng-etal-2024-multimodal, wang2024mfc}. Explanation-based frameworks~\cite{shu2019defend,zeng2023multi,ying2023bootstrapping,qi2024sniffer,lu2025dammfnd}
produce localized post-hoc justifications yet lack an explicit reasoning layer connecting low-level cues to high-level veracity.  
Our framework addresses this gap by bridging low-level multimodal features with interpretable, concept-driven veracity prediction.

\textbf{Concept Models and Structured Reasoning.}
Concept Bottleneck Models (CBMs)~\cite{koh2020concept} predict human-interpretable concepts as intermediates before task outputs, allowing user intervention and transparent explanations. However, their fixed and flat concept space limits scalability for complex reasoning tasks like MMD. Neuro-symbolic models integrate structured reasoning with neural representations, combining the flexibility of deep learning with the interpretability of symbolic inference, particularly in visual question answering and scene understanding~\cite{vedantam2019probabilistic}. Our work shares this modular reasoning spirit but extends it to a \emph{learnable concept graph} that evolves with new manipulation patterns.
Graph-of-Thought (GoT) prompting~\cite{10.1609/aaai.v38i16.29720} generalizes Chain-of-Thought (CoT) reasoning~\cite{cao2026diffcot} to graph-structured inference in large languge models (LLMs), enabling dynamic reasoning branches. In contrast, PCGR embeds a probabilistic concept graph directly within the model architecture, enabling structured multimodal reasoning without external prompting.

%% file: sec/3_problemform.tex
\section{Problem Formulation}
\label{sec:prob}
We define a multimodal misinformation dataset as $\{\mathcal{N}\}$, where each instance $\mathcal{N} = (X, I, y)$ consists of a text $X$, an accompanying image $I$, and a binary label $y \in \{0, 1\}$, where $1$ denotes \textit{fake} and $0$ denotes \textit{real}. 
We construct a \emph{growing} concept set $\mathcal{C} = \{c_k\}_{k=1}^K$, organized as a layered directed acyclic graph (DAG) $\{\mathcal{L}_r\}_{r=0}^R$, where each layer $\mathcal{L}_r$ contains semantically coherent concepts, and directed edges connect $\mathcal{L}_r$ to $\mathcal{L}_{r+1}$. Each concept $c_k$ is expressed in human-readable text.   

The model first predicts a probability $p_k$ for each concept via  
$f: (X, I) \rightarrow (p_1, p_2, \dots, p_K)$,  
where $q_k$ represents the likelihood that the multimodal input supports concept $c_k$.  
The predicted concepts $\{c_k\}$ and their probabilities $\{p_k\}$ are then translated into a concise, human-readable explanation that summarizes the reasoning process, forming an evidence-based narrative of why the input is classified as fake or real.  
Next, a veracity classifier aggregates the concept probabilities to infer the final truth label via  
$g: (p_1, p_2, \dots, p_K) \rightarrow y$.  
The DAG structure enables top-down reasoning, where higher-level abstract concepts in $\mathcal{L}_{r+1}$ condition more specific lower-level concepts in $\mathcal{L}_r$.  
This formulation supports structured and interpretable reasoning: concept probabilities act as intermediate evidence nodes, while the layered graph structure enhances transparency and adaptability to evolving misinformation patterns.


When fine-grained labels (e.g., \textit{textual distortion}, \textit{visual distortion}, \textit{cross-modal inconsistency}, \textit{real}) are available, they are used as \emph{anchor concepts} in the first layer $\mathcal{L}_0$, with their probabilities directly reported.  
In the absence of such labels, all concept layers are generated automatically through the model’s concept growth process.
 
%

%% file: sec/4_method.tex
\section{Methodology}
\begin{figure*}[t]
    \centering
    \includegraphics[width=0.98\linewidth]{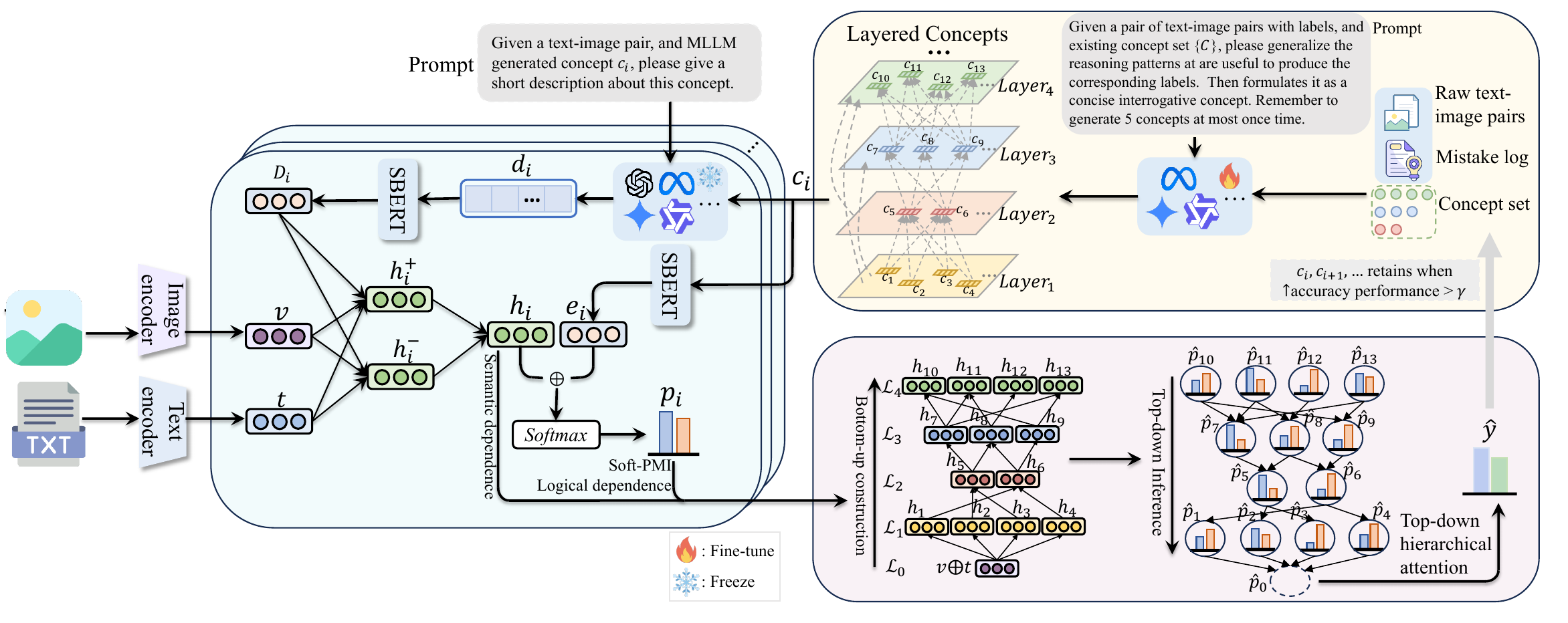}
    \vspace{-0.4cm}
    \caption{The Probabilistic Concept Graph Reasoning (PCGR) framework. PCGR organizes multimodal evidence into interpretable concept nodes and performs hierarchical probabilistic reasoning to infer claim veracity. }
    \vspace{-0.4cm}
    \label{fig:model}
\end{figure*}
\label{sec:method}
Multimodal misinformation evolves rapidly through new manipulation tactics, rendering fixed taxonomies brittle and monolithic end-to-end models difficult to adapt. An effective detector should therefore be modular, probabilistic, and incrementally extensible. To this end, we propose Probabilistic Concept Graph Reasoning (PCGR) shown in Figure~\ref{fig:model}, a framework that (i) automatically induces layered concepts to capture emerging manipulation patterns (in yellow), (ii) performs soft probabilistic reasoning over inter-concept dependencies (in blue), and (iii) employs hierarchical attention to support both coarse- and fine-grained predictions (in pink).

\subsection{Automatic Concept Growth}\label{sec:conceptgrowing}
PCGR adapts to new manipulation tactics by continuously discovering and integrating new concepts. As shown in Figure~\ref{fig:model} (yellow part), each \textit{growth round} includes concept proposal via an MLLM (e.g, Qwen3-omni) prompting and automated filtering, repeated up to six rounds with at most five new concepts per round. 
\textbf{Seed Selection and MLLM Prompting.}
To identify weaknesses, we maintain a \textit{mistake log} of high-loss samples, updated each round. Representative text-image seed pairs are selected via k-means clustering in embedding space and fed into the MLLM using the template prompt shown in Figure ~\ref{fig:model}. 
These seeds are used to prompt an MLLM (e.g., GPT-5) to generate abstract, reusable concept candidates. Acting as an expert fact-checker, the MLLM (1) analyzes why each sample is misleading, (2) generalizes the reasoning pattern, i.e., identifying reusable diagnostic phrases, such as ``Exaggerated emoji'', for forming interrogative concepts, and (3) formulates it as a concise interrogative concept, e.g., ``Does the text exaggerate the event?'' or ``Is the image contextually inconsistent with the caption?''.



\textbf{Concept Filtering and Validation.}
MLLM-proposed concepts are screened by a three-stage filter to ensure quality, novelty, and usefulness before integration. A concept is kept only if it satisfies all of the following three conditions:
\textbf{1) Semantic Uniqueness:} Prevent paraphrases (e.g., ``fake image'' vs. ``manipulated picture'') by requiring low similarity to any existing concept using $\cos(c_{\text{new}}, c_{\text{old}}) \le 0.8$. \textbf{2) Statistical Independence:} Avoid functionally redundant concepts by enforcing Pearson's correlation $|\mathrm{corr}(p_{\text{new}}, p_{\text{old}})| \le 0.9$ on validation predictions.  \textbf{3) Informative Activation:} Retain concepts that are neither trivial nor degenerate based on $\mathbb{E}[p_{\text{new}}] \in [0.05, 0.95]$ using validation data.
Concepts passing all checks are added to the graph as \emph{soft predicates}, enabling continuous, partially satisfied constraints in our probabilistic reasoning. A detailed Algorithm~\ref{algo:conceptgrow} can be found in the supplementary material. We then prompt MLLM to generate a short description $D_i$ for the text-image pair under the guidance of filtered concept $c_i$.

\subsection{Text-Image Encoding}
Given a text-image pair $(X,I)$, we use two independent multimodal encoders (e.g., CLIP~\cite{radford2021learning})
to preserve modality-specific features and reduce alignment bias. We extract text and image embeddings as $t=\text{CLIP}(X;\theta_X)$ and $v=\text{CLIP}(I;\theta_I)$. For each concept's description $D_i$, we extract a textual embedding $d_i$ with Sentence-BERT~\cite{reimers2019sentence}.

To model uncertainty, where the absence of evidence does not imply falsity, we represent each concept by positive and negative prototypes $h_i^+$ and $h_i^-$, denoting its active and inactive states:
\begin{equation}
\small
    h_i^+=\varphi_i^+\cdot(v\oplus t\oplus d_i), \quad 
    h_i^-=\varphi_i^-\cdot(v\oplus t\oplus d_i),
\end{equation}
where ${\varphi}^+$ and ${\varphi}^-$ are model's parameter matrices learned to map multimodal evidence into active ($h^+$, e.g., manipulation artifacts) and inactive ($h^-$, e.g., natural features) prototype states, respectively. The final representation combines both prototypes:
\begin{equation}\label{equ:conceptrep}
\small
    h_i = \tau_i h_i^+ + (1-\tau_i)h_i^-,
\end{equation}
with $\tau_i$ as a learnable weight.

\subsection{Concept Probability Computation}
For each concept $c_k$, we obtain a textual embedding with Sentence-BERT~\cite{reimers2019sentence} and condition a lightweight multimodal interaction to estimate $p_k$:
\begin{equation} \label{equ:conceptprobability}
\small
\begin{split}
    e_k &= \text{SBERT}(c_k; \theta_k), \\
    \ell_k &= h_k \; \oplus \; \mu_k\, U^\top \!\diag(\phi(e_k))\, V^\top \nu_k, \\
    p_k &= \mathrm{Linear}(w_k \ell_k + b_k),
\end{split}
\end{equation}
where $U,V \in \mathbb{R}^{d\times r}$ ($r \!\ll\! d$) are shared low-rank projections, $\phi(\cdot)$ is an MLP, and $\mu_k,\nu_k,w_k,b_k$ are learnable parameters. $\diag(\phi(e_k))$ is implemented via element-wise products for efficiency.


\subsection{Concept Graph Construction}

We place each image-text instance at the bottom layer $\mathcal{L}_0$ and grow higher layers bottom-up from residual errors. Inference is top-down, i.e., abstract hypotheses in $\mathcal{L}_r$ provide some priors to refine evidence in $\mathcal{L}_{r+1}$. \looseness=-1

An ideal concept graph should capture both compositional and inferential relationships between concepts. Intuitively, it should allow lower-level concepts to combine into higher-level abstractions, thereby modeling subtle yet powerful reasoning patterns. 
To formalize this, we score a candidate edge between $i\!\in\!\mathcal{L}_r$ and $j\!\in\!\mathcal{L}_{r+1}$ by combining semantic, statistical\footnote{Statistical dependence is estimated using soft Pointwise Mutual Information (Soft-PMI) computed over concept probabilities within each training batch.}, and logical signals:
\begin{equation}\label{equ:pre-score}
\small
\nonumber
  \begin{split}
      & s_{ij} = 
      - \underbrace{\alpha \cos(h_i,h_j)}_{\text{Semantic Dependence}}
      + \underbrace{\beta \log\frac{\bar p_{ij}}{\bar p_i\,\bar p_j}}_{\text{Soft-PMI}}
      + \underbrace{\gamma r^{\text{ent}}_{ij} - \delta r^{\text{contr}}_{ij}}_{\text{Logical Dependence}}, \\
      & \bar p_i = \tfrac{1}{B}\sum_b p_i^{(b)}, \quad
      \bar p_{ij} = \tfrac{1}{B}\sum_b p_i^{(b)} p_j^{(b)}, \\
      & r^{\text{ent}}_{ij}, r^{\text{neutr}}_{ij}, r^{\text{contr}}_{ij} = \text{NLI}(h_i, h_j),
  \end{split}  
\end{equation}
where $t_i$ and $t_j$ denote the embeddings of concepts $i$ and $j$, $p_i^{(b)}, p_j^{(b)}$ are their predicted probabilities for batch $b$. The weights $\alpha$, $\beta$, $\gamma$, and $\delta$ control the relative contributions. We employ DeBERTa-v3-Large-MNLI~\cite{debertlarge} for NLI to estimate entailment and contradiction scores between concept pairs. 

Finally, edges between $i \in \mathcal{L}_r$ and $j \in \mathcal{L}_{r+1}$ are created when $ s_{ij} > \zeta$, where $\zeta$ is a predefined threshold set to 0.55.
A larger $s_{ij}$ indicates that $j$ is a meaningful abstraction or composition of $i$. 

\subsection{Multimodal Misinformation Detection}
Higher-level concepts in $\mathcal{L}_{r+1}$ act as conditional priors over lower-level refinements in $\mathcal{L}_r$. 
We realize this with top-down hierarchical attention. For $i \in \mathcal{L}_{r}$ and $\text{Pa}(i)\subseteq \mathcal{L}_{r+1}$, we define:
\begin{equation}\label{equ:attention}
\small
    \alpha_{ij} = 
    \frac{\exp(t_i \cdot t_j)}{\sum_{j' \in \text{Pa}(i)} \exp(t_i \cdot t_{j'})},
\end{equation}
with $t_i,t_j$ as the concept embeddings. 

Since veracity assessment relies on multiple consistency cues that must hold simultaneously, the aggregation mechanism should approximate a logical \textsc{AND}. We adopt a hierarchical probabilistic aggregation scheme whose multiplicative form enhances robustness to noise, maintains probabilistic interpretability, and yields better calibration than additive or voting approaches:
\begin{equation} \label{equ:finaldecision}
\small
\begin{split}
    \hat{p}_i &= \lambda p_i \cdot (1-\lambda) \prod_{j \in \text{Pa}(i)} (\alpha_{ij} p_j), \\
    \hat y &= \hat{p}_0, 
\end{split}    
\end{equation}
where $\lambda$ is a confusion parameter, $p_i$ is the probability of concept $i$ (from {Eq.~\ref{equ:conceptprobability}}), $\text{Pa}(i)$ denotes its parents nodes, and $\alpha_{ij}$ is the attention weight defined in Eq.~\ref{equ:attention}.



\subsection{Model Training}
Our model is trained end-to-end through alternating optimization between the concept generation and misinformation detection modules. The overall objective combines veracity prediction with a concept orthogonality regularizer:
\begin{equation}
\small
L = (1 - \eta)L_{\text{veracity}} + \eta L_{\text{ortho}},
\end{equation}
where $L_{\text{veracity}}$ is the detection loss, $L_{\text{ortho}}$ enforces independence among concept embeddings, and $\eta$ balances the two terms. The losses are defined as:
\begin{equation}
\small
\nonumber
    \begin{split}
        L_{\text{veracity}} &= -\sum_{n=1}^{N} \big[y_n \log \hat{y}_n + (1-y_n) \log(1-\hat{y}_n)\big], \\
        L_{\text{ortho}} &= \sum_{i \neq j} \frac{q_i^\top \cdot q_j}{\|q_i\|_2^2 \cdot \|q_j\|_2^2}, 
    \end{split}
\end{equation}
where $N$ is the number of samples, $\hat{y}_n$ the prediction for sample $n$, $y_n$ its ground truth, and $q_i$, $q_j$ the embeddings of concepts $i$ and $j$.

\textbf{Optional Anchors.}
When fine-grained labels (e.g., \emph{textual distortion}, \emph{visual distortion}, \emph{cross-modal inconsistency}, \emph{true}) avail, they are used as anchor concepts in $\mathcal{L}_0$ and supervised via Eq.\ref{equ:conceptprobability}. Higher-level layers $\mathcal{L}_1,\mathcal{L}_2,\dots$ then grow automatically. Alongside the overall misinformation score $\hat{y}$, the model outputs anchor probabilities $\{\mu_{\text{text}}, \mu_{\text{vis}}, \mu_{\text{cross}}\}$ and enforces consistency between coarse veracity and fine-grained labels.



%% file: sec/5_experiment.tex
\section{Experiments and Results}
\label{sec:experiment}

\subsection{Experimental Setup}

\begin{figure}[htbp]
    \centering
    \includegraphics[width=0.95\linewidth, trim=0 0 0 0, clip]{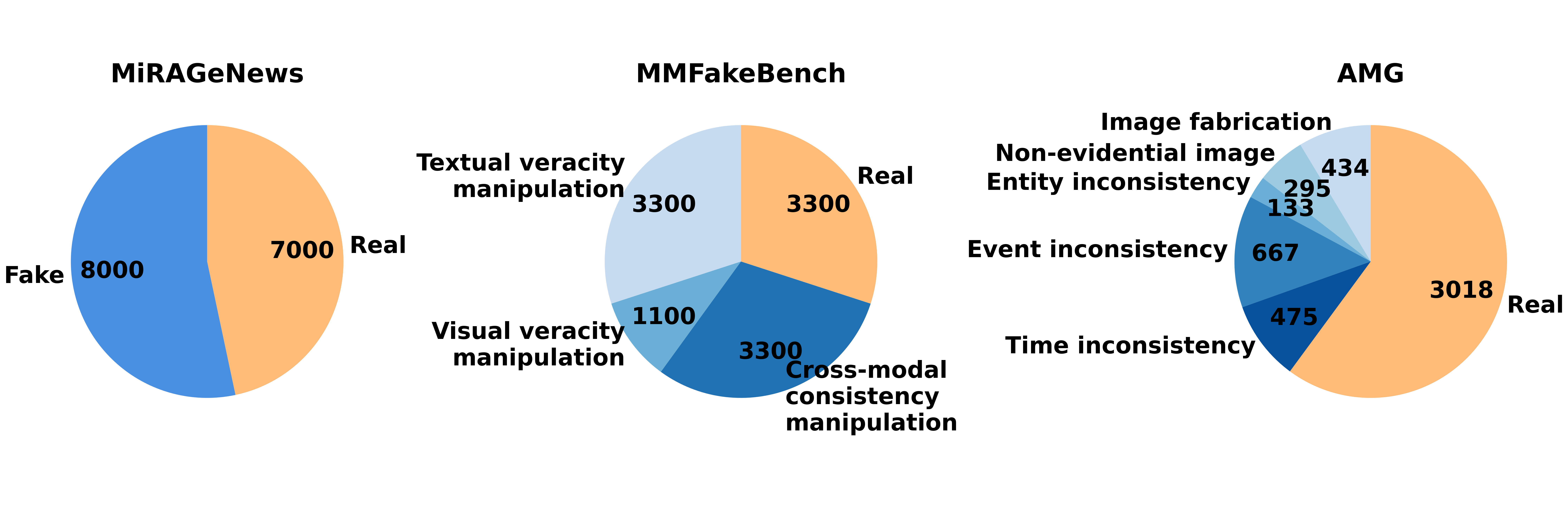}
    \vspace{-0.6cm}
    \caption{Statistics of three benchmark datasets.}
    \vspace{-0.2cm}
    \label{fig:pie}
\end{figure}

\paragraph{Datasets.} We evaluate PCGR on three challenging multimodal misinformation detection benchmarks.
\textbf{MiRAGeNews}~\cite{huang2024miragenews}: A binary dataset targeting AI-generated news. Its test set includes content from unseen image generators and publishers, serving as a rigorous out-of-domain (OOD) benchmark for robustness and generalization.
\textbf{MMFakeBench}~\cite{liummfakebench}: A large-scale benchmark covering both coarse- and fine-grained multimodal misinformation. It includes binary labels {\emph{fake, real}} and four fine-grained subtypes {\emph{textual veracity manipulation}, \emph{visual veracity manipulation}, \emph{cross-modal consistency manipulation}, \emph{real}}.
\textbf{AMG}~\cite{guo2025each}: A multi-platform dataset annotated with binary veracity labels and six fine-grained types (\emph{image fabrication}, \emph{non-evidential image}, \emph{entity/event/time inconsistency}, \emph{true}). Figure~\ref{fig:pie} summarizes their statistics.
\vspace{-0.3cm}

\begin{figure*}[htbp]
    \centering
    \includegraphics[width=0.98\linewidth, trim=0 0 0 0, clip]{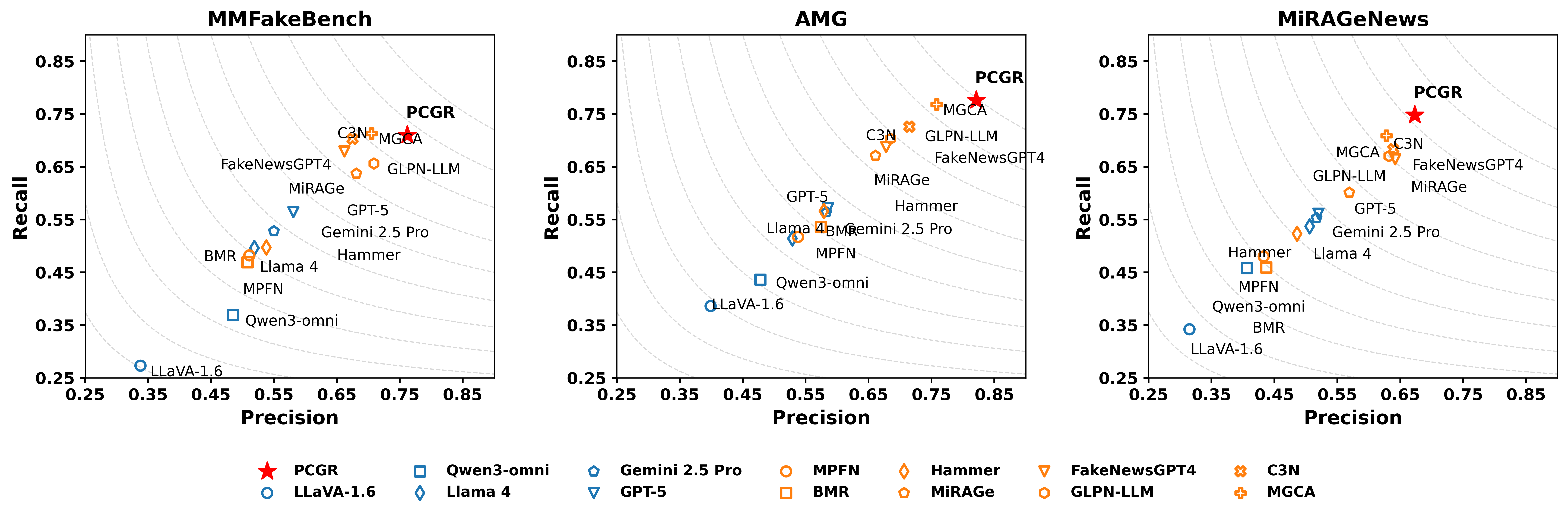}
    \vspace{-0.4cm}
    \caption{Precision-Recall curves for MMD results on MMFakeBench and AMG (in-domain) and MiRAGeNews (OOD). Blue and green markers denote general-purpose MLLMs and task-specific detectors, respectively. Dashed gray lines indicate iso-F1 contours.}
    \vspace{-0.3cm}
    \label{fig:pr_curves}
\end{figure*}

\textbf{Baselines.} We compare PCGR against 13 strong baselines, grouped into two categories (details in the supplementary):
(1) \textbf{General MLLMs:} Multimodal large language models designed for broad vision-language tasks, including LLaVA-1.6~\cite{liu2024llavanext}, Qwen3-Omni~\cite{xu2025qwen3}, Llama 4~\cite{llama4}, Gemini 2.5 Pro~\cite{comanici2025gemini}, and GPT-5~\cite{gpt5}.
    (2) \textbf{Multimodal Detectors:} Specialized models for misinformation detection, including MPFN~\cite{jing2023multimodal}, BMR~\cite{ying2023bootstrapping}, HAMMER~\cite{shao2023detecting}, MiRAGe~\cite{huang-etal-2024-miragenews}, FKA-Owl~\cite{liu2024fka}, GLPN-LLM~\cite{hu-etal-2025-synergizing}, C3N~\cite{qiao2025improving}, MGCA~\cite{guo2025each}.
For all baselines, we adopt the hyperparameters recommended in their original papers and fine-tune key settings (e.g., learning rate, embedding size). We implement baselines and PCGR with PyTorch~\cite{paszke2019pytorch}. More implementation details are given in the supplementary material.

\textbf{Metrics.} For coarse-level detection, we report Accuracy (Acc), Precision (Pre), Recall (Rec), and F1-Score (F1). For fine-grained detection on \textsc{MMFakeBench} and \textsc{AMG}, we report Micro-F1 (Mic-F1) and Macro-F1 (Mac-F1). Implementation details are presented in Supplementary~\ref{supp:imple}.

\begin{table}[t!]
\small
\centering
\caption{Coarse-level MMD accuracy (\%) and F1 across three benchmarks.}
\vspace{-0.3cm}
\resizebox{0.98\columnwidth}{!}{
\begin{tabular}{l|cc|cc|cc}
\toprule
\multirow{2}{*}{\textbf{Method}}
& \multicolumn{2}{c|}{\textbf{MiRAGeNews}}
& \multicolumn{2}{c|}{\textbf{MMFakeBench}}
& \multicolumn{2}{c}{\textbf{AMG}} \\
\cmidrule{2-7}
& Acc & F1 & Acc & F1 & Acc & F1 \\
\midrule
\multicolumn{7}{l}{\textit{General MLLMs}} \\
\midrule
LLaVA-1.6       & 31.9 & 32.8 & 31.2 & 30.2 & 37.5 & 39.2 \\
Qwen3-omni      & 30.2 & 43.0 & 33.9 & 41.9 & 39.2 & 45.6 \\
Llama 4         & 43.1 & 52.1 & 46.8 & 50.7 & 48.8 & 52.1 \\
Gemini 2.5 Pro  & 52.9 & 53.4 & 50.3 & 53.9 & 56.8 & 57.3 \\
GPT-5           & 56.8 & 54.0 & 58.8 & 57.2 & 59.9 & 57.9 \\
\midrule
\multicolumn{7}{l}{\textit{Multimodal Detectors}} \\
\midrule
MPFN            & 49.9 & 45.5 & 53.4 & 49.6 & 55.4 & 52.7 \\
BMR             & 49.5 & 44.8 & 52.3 & 48.8 & 57.3 & 55.4 \\
Hammer          & 52.8 & 50.4 & 55.7 & 51.7 & 60.1 & 57.2 \\
MiRAGe          & 61.1 & 58.5 & 64.8 & 65.8 & 69.2 & 67.1 \\
FKA-Owl    & 63.7 & 65.3 & 64.7 & 67.9 & 70.7 & 68.7 \\
GLPN-LLM        & 66.2 & 65.0 & 69.3 & 68.1 & 73.6 & 70.3 \\
C3N             & 70.4 & 66.0 & 73.6 & 70.3 & 75.3 & 72.6 \\
MGCA            & 72.3 & 66.6 & 74.1 & 71.3 & 78.2 & 76.8 \\
\midrule
\textbf{PCGR}   & \textbf{80.2} & \textbf{70.9} & \textbf{80.6} & \textbf{73.5} & \textbf{84.3} & \textbf{79.8} \\
\bottomrule
\end{tabular}}
\label{tab:Coarse-levelResult}
\vspace{-0.5cm}
\end{table}

\subsection{Coarse-level MMD Results}
As shown in Table~\ref{tab:Coarse-levelResult} and Figure~\ref{fig:pr_curves}, PCGR consistently outperforms all 13 baselines across all datasets and metrics. Compared to the strongest baseline (MGCA), PCGR achieves accuracy gains of 7.9\%, 6.5\%, and 6.1\% on \textsc{MiRAGeNews}, \textsc{MMFakeBench}, and \textsc{AMG}, respectively. These improvements demonstrate that probabilistic concept graph reasoning effectively enhances multimodal misinformation detection.
General-purpose MLLMs perform substantially worse than dedicated multimodal detectors, highlighting the importance of explicit cross-modal fusion for misinformation verification. GPT-5 performs relatively better among general MLLMs, likely due to its broader pretraining coverage up to September 2024.
Among task-specific detectors, joint modeling approaches such as MGCA achieve stronger performance than isolated classifiers, confirming the benefit of multi-granular supervision. Notably, most baselines show degraded performance on the OOD dataset \textsc{MiRAGeNews}, while PCGR maintains stable accuracy across domains, demonstrating superior robustness and generalization.

\begin{table}[t!]
  \centering
  \small
  \vspace{-0.1cm}
  \caption{Fine-grained detection performance on \textsc{MMFakeBench} (4 classes) and \textsc{AMG} (6 classes).}
  \vspace{-0.3cm}
  \resizebox{0.45\textwidth}{!}{
    \begin{tabular}{l|rr|rr}
    \toprule
          & \multicolumn{2}{c|}{MMFakeBench} & \multicolumn{2}{c}{AMG} \\
    \midrule
          & \multicolumn{1}{l}{Mic-F1} & \multicolumn{1}{l|}{Mac-F1} & \multicolumn{1}{l}{Mic-F1} & \multicolumn{1}{l}{Mac-F1} \\
    \midrule
    LLaVA-1.6 & 24.9  & 24.5  & 28.6  & 25.7 \\
    Qwen3-omni & 25.6  & 22.7  & 30.6  & 27.3 \\
    Llama 4 & 28.7  & 25.8  & 35.3  & 30.6 \\
    Gemini 2.5 Pro & 46.1  & 48.3  & 53.8  & 51.4 \\
    GPT - 5 & 55.3  & 54.9  & 60.7  & 55.5 \\
    MGCA  & 60.3  & 55.1  & 72.8  & 57.2 \\
    \midrule
    \textbf{PCGR} & \textbf{68.6} & \textbf{56.9} & \textbf{75.6} & \textbf{59.9} \\
    \bottomrule
    \end{tabular}%
    }
  \label{tab:Fine-levelResults}%
  \vspace{-0.5cm}
\end{table}%

\subsection{Fine-grained Detection Results}
Since most multimodal detectors cannot identify fine-grained manipulation types (except MGCA), we compare PCGR only with MGCA and general-purpose MLLMs. As \textsc{MiRAGeNews} lacks fine-grained manipulation labels, we evaluate on \textsc{MMFakeBench} (4 manipulation types) and \textsc{AMG} (6 manipulation types). Fine-grained categories are used as concept anchors in the first layer of the concept graph.
As shown in Table~\ref{tab:Fine-levelResults}, PCGR achieves the best performance across both datasets, confirming its ability to integrate and align with human-defined expert concept sets. Its superior accuracy in detecting diverse manipulation types highlights the model’s flexibility in supporting both coarse- and fine-grained reasoning. MGCA performs better than general MLLMs because its joint modeling framework allows mutual reinforcement between coarse- and fine-grained features.


\subsection{Ablation Study}
To assess the contribution of each major component in PCGR, we perform an ablation study on the \textsc{AMG} dataset using the following variants: 1) \textbf{w/o ma}: replace multiplicative aggregation with voting. 2) \textbf{w/o acg}: remove automatic concept generation and rely only on the initial MLLM-generated concepts; 3) \textbf{w/o dag}: replace the layered DAG with a flat structure and average all concept probabilities; 4) \textbf{w/o hat}: replace the top-down hierarchical attention with standard dot-product attention; 5) \textbf{w/o alt}: remove the alternating training schedule; 6) \textbf{w/o warm}: remove the warm-up phase in concept growth; 7) \textbf{w/o cf}: disable the concept filtering and validation process.  


\begin{figure}[t!]
    \centering
    \includegraphics[width=0.92\linewidth, trim=0 0 0 0, clip]{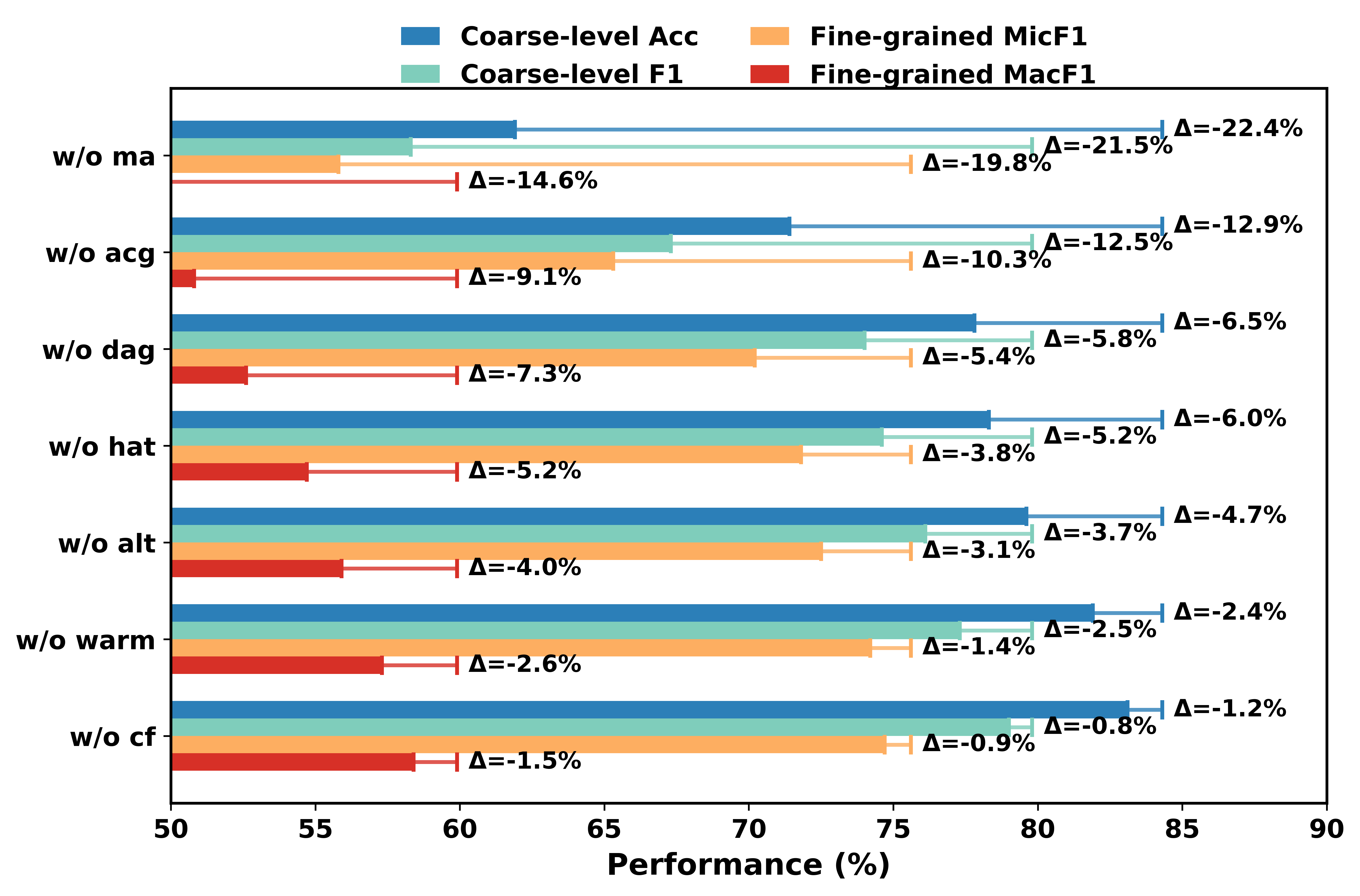}
    \vspace{-0.4cm}
    \caption{Ablation results for coarse-level and fine-grained  detection on the AMG dataset.}
    \vspace{-0.62cm}
    \label{fig:ablation}
\end{figure}

As shown in Figure~\ref{fig:ablation}, removing any major component leads to noticeable drops in both coarse- and fine-grained performance, confirming the contribution of each design choice.
The largest decline occurs with \textbf{w/o acg}, where Mic-F1 and Mac-F1 drop by about 12.9\% and 12.5\%, respectively, underscoring the critical role of automatic concept generation.
Both \textbf{w/o dag} and \textbf{w/o hat} degrade performance significantly, validating the effectiveness of PCGR’s \emph{build bottom-up, infer top-down} paradigm.
\textbf{w/o alt} shows that alternating training prevents the model from converging to local minima, while \textbf{w/o warm} and \textbf{w/o cf} highlight the benefits of warm-up initialization and concept-quality filtering in stabilizing training and improving final accuracy.

\subsection{Case Study}

\begin{figure}
    \centering
    \includegraphics[width=0.5\textwidth]{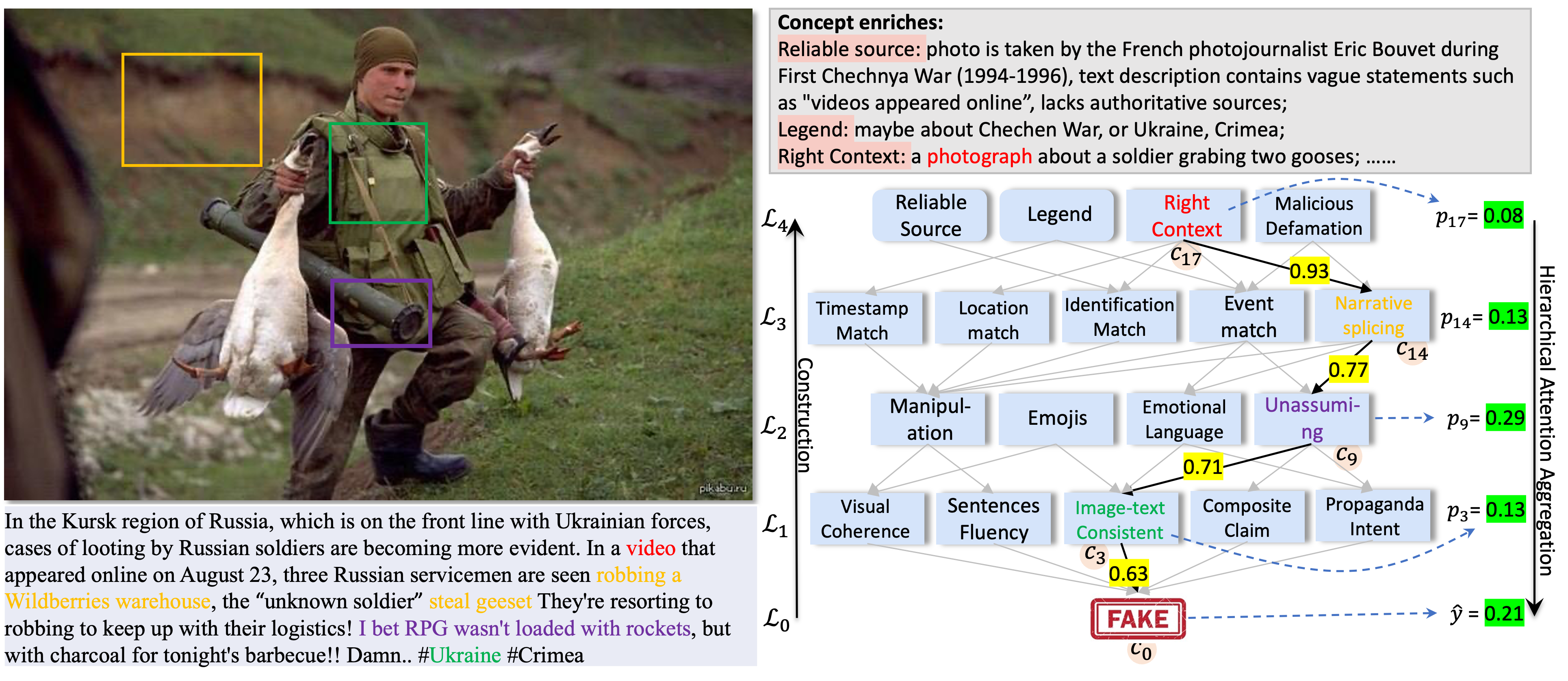}
    \vspace{-0.6cm}
    \caption{A case study illustrating the hierarchical probabilistic aggregation process.}
    \vspace{-0.3cm}
    \label{fig:casestudy}
\end{figure}

To illustrate how PCGR performs hierarchical probabilistic reasoning, we visualize a text-image pair from the AMG dataset that our model correctly classifies as \emph{fake}. Figure~\ref{fig:casestudy} shows the input (left) and the induced concepts with their dependencies (right).
Concept $c_{17}$ (``right context’’) captures a key inconsistency: the image shows a \emph{photograph}, while the text refers to a \emph{video}. This mismatch yields a low concept probability (0.08). Each blue box denotes a concept in the layered graph ($\mathcal{L}_0$–$\mathcal{L}_4$), with yellow highlights indicating the dominant parent contributing to each node.
The most influential node for the final decision is $c_3$ at layer $\mathcal{L}_1$, whose strongest parent is $c_9$ in $\mathcal{L}_4$, which in turn is influenced by $c_{17}$. During top-down aggregation, $c_{17}$ contributes to $c_9$ via attention weight $\alpha_{17\to 9}=0.93$, propagating a value of $0.08\times 0.93=0.0774$. This recursive process continues until the root node ($c_0$) yields the final veracity score $\hat{y}=0.21$, which PCGR labels as \emph{fake}.
\vspace{-0.3cm}

%% file: sec/6_conclusion.tex
\section{Conclusion}
\label{sec:conclu}
We propose PCGR, a probabilistic concept-graph reasoning framework for multimodal misinformation detection equipped with an automatic concept-growth mechanism. PCGR continually discovers and integrates new reasoning concepts via MLLMs, enabling adaptation to emerging manipulation tactics. A layered probabilistic concept graph reframes detection as soft, hierarchical reasoning, while a top-down attention mechanism aggregates uncertain concept-level signals across layers to support both coarse- and fine-grained detection.
Across multiple datasets and domains, PCGR consistently surpasses strong baselines at both coarse- and fine-grained level tasks.


%% file: sec/X_suppl.tex
\clearpage
\setcounter{page}{1}
\maketitlesupplementary
\appendix

\section{Algorithm for Concept Growth}

Algorithm~\ref{algo:conceptgrow} summarizes the alternating procedure and stops early after two non-improving rounds (typically $\#rounds \le 6 $).

\begin{algorithm}[htbp]
\small
\caption{Alternating Concept Growth (per round)}
\begin{algorithmic}[1]
\State \textbf{Input:} concept graph $(\mathcal{C}, \{\mathcal{L}_r\})$, encoders, train set $\mathrm{Tr}$, validation set $\mathrm{Va}$
\State Propose up to 5 new concepts via LLM prompting (\S\ref{sec:conceptgrowing})
\State Initialize new edges $\mathcal{L}_{r}\!\to\!\mathcal{L}_{r+1}$ via Eq.~\ref{equ:conceptprobability} and Eq.~\ref{equ:pre-score}
\State \textbf{Warm-up:} train concept classifier for 2 epochs; freeze previous layers
\State \textbf{Joint training:} unfreeze current and preceding layers for 6 epochs
\State \textbf{Validation check:} retain round only if $\Delta$NLL $\ge \eta$ and $\Delta$AUC is significant
\State Consolidate valid concepts (3 epochs), save checkpoint
\State \textbf{Return:} concept graph $(\mathcal{C}, \{\mathcal{L}_{r+1}\})$
\end{algorithmic}
\label{algo:conceptgrow}
\end{algorithm}

\section{Experiment Baselines}

\textbf{General MLLMs}:
We use the following general purpose MLLMs to do zero-shot misinformation detection in our paper.
\begin{itemize}
    \item \textbf{LLaVA-1.6}~\cite{liu2024llavanext}: A large multimodal model that combines a vision encoder and Vicuna for general-purpose visual and language understanding.
    \item \textbf{Qwen3-omni}~\cite{xu2025qwen3}: A large multimodal model capable of processing multiple modalities, and generating real-time text or speech response.
    \item \textbf{Llama 4}~\cite{llama4}: A natively multimodal that enable text and multimodal experiences.
    \item \textbf{Gemini 2.5 Pro}~\cite{comanici2025gemini}: A multimodal reasoning model aims to solve complex problems.
    \item \textbf{GPT-5}~\cite{gpt5}: A deeper reasoning multimodal model for harder problems such as writing, research, analysis, and so on.
   
\end{itemize}

\textbf{Multimodal Detectors}:

\begin{itemize}
    \item \textbf{MPFN}~\cite{jing2023multimodal}: A multimodal detector captures and fuses both shallow and deep level information text and images for misinformation detection.
    \item \textbf{BMR}~\cite{ying2023bootstrapping}: A multimodal detector 
    that bootstrap each modality's representation by intial predictions for single modality to get the final misinformation detection result.
    \item \textbf{HAMMER}~\cite{shao2023detecting}: A multimodal detector utilizes the fine-grained interaction between different modalities for mistinformation detection. 
    \item \textbf{MiRAGe}~\cite{huang-etal-2024-miragenews}: A simple multimodal detector that fuses text- and image-level concept bottleneck models' results to combat the spread of AI-generated misinformation.
    \item \textbf{FKA-Owl}~\cite{liu2024fka}: A multmodal detector that leverages forgery-specific knowledge to augment MLLMs to reason about manipulations.
    \item \textbf{GLPN-LLM}~\cite{hu-etal-2025-synergizing}: A multimodal detector that integrates LLM capabilities via label propagation techniques to enhance prediction accuracy. 
    \item \textbf{C3N}~\cite{qiao2025improving}: A multimodal detector that captures learnable patterns of cross-modal content correlations to facilitate news classification.
    \item \textbf{MGCA}~\cite{guo2025each}: A multimodal detector that aligns multi-granularity clues for both misinformation and distortion detection.
\end{itemize}

\section{Implementation Details}\label{supp:imple}
We implement PCGR with PyTorch~\cite{paszke2019pytorch} and trained it on four NVIDIA A100 GPUs. Each dataset was split into 70\% training, 20\% validation, and 10\% testing. The learning rate was set to $1 \times 10^{-4}$ and batch size to 32. Concept generation was performed using the GPT-5 API. All parameters were optimized using the Adam optimizer~\cite{collobert2011natural}, and training stopped upon convergence or after 150 epochs. For automatic concept growth, MLLM prompting occurs only during training and is never used at inference. And each round queries only the previous round’s mistake log, which shrinks over rounds. And Table~\ref{tab:TrainingTime} shows PCGR has comparable training/inference time to MGCA on 4×A100 GPUs.
\begin{table}[htbp]
  \centering
  \vspace{-0.3cm}
  \caption{Training and inference time of MGCA and PCGR.}
  \vspace{-0.3cm}
  \resizebox{1\columnwidth}{!}{
    \begin{tabular}{l|lll|lll}
    \toprule
    \multicolumn{1}{c|}{\multirow{2}[4]{*}{Method}} & \multicolumn{3}{c|}{Training time} & \multicolumn{3}{c}{Inference time} \\
\cmidrule{2-7}          & MiRAGeNews & MMFakeBench & AMG   & MiRAGeNews & MMFakeBench & AMG \\
    \midrule
    MGCA  & 193mins & 300mins & 180mins & 20mins & 33mins & 18mins \\
    \midrule
    \textbf{PCGR} & \textbf{220mins} & \textbf{350mins} & \textbf{178mins} & \textbf{20mins} & \textbf{32mins} & \textbf{20mins} \\
    \bottomrule
    \end{tabular}
    \vspace{-0.3cm}
    }
  \label{tab:TrainingTime}%
\end{table}%

\section{Fine-grained Detection Results}
\begin{figure}
    \centering
    \includegraphics[width=1.1\columnwidth]{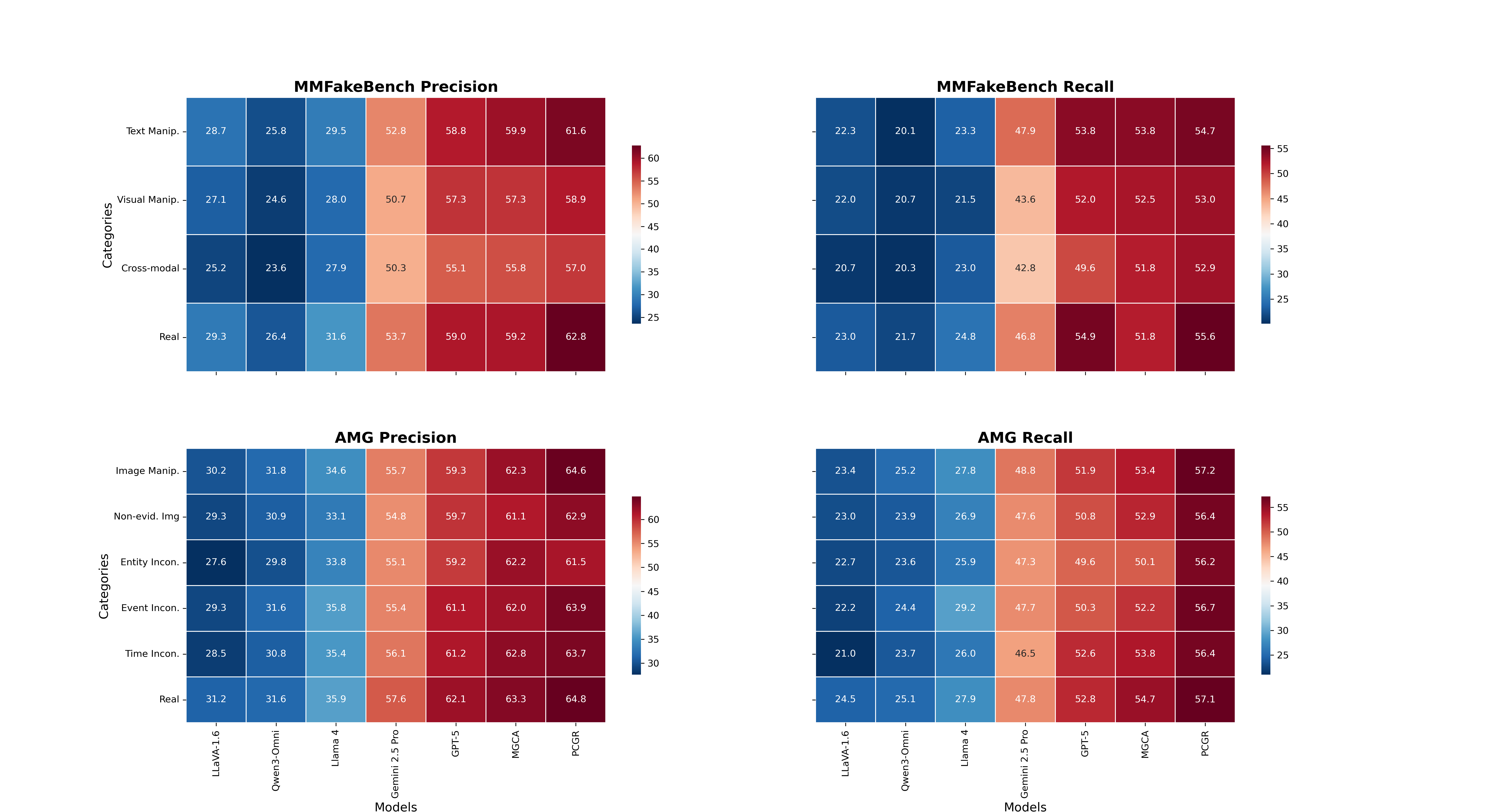}
    \caption{Precision and recall heatmap for fine-grained level detector results on MMFakeBench and AMG datasets. Score magnitudes are chromatically encoded, ranging from blue (lowest) to red (highest).}
    \label{fig:fine-grainedprerec}
\end{figure}

Figure~\ref{fig:fine-grainedprerec} shows the precision and recall for each subcategory in MMFakeBench and AMG datasets. We can observe that PCGR beats all baselines across two datasets, demonstrating the superior performance of our model for fine-grained classification tasks.

\section{Sensitivity Study}\label{sect:sensi}

\begin{figure}
    \centering
    \includegraphics[width=1\linewidth]{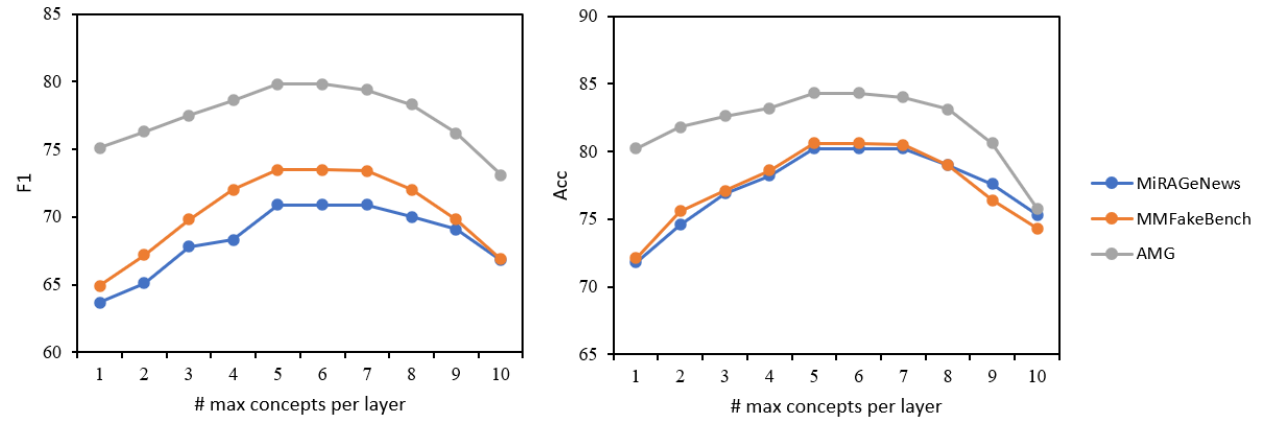}
    \caption{Impact of the max concepts constraint per layer.}
    \label{fig:conceptsense}
\end{figure}

We conduct a sensitivity study to investigate the impact of the maximum number of concepts per layer. Specifically, we evaluate the F1 and accuracy scores of our PCGR model on the MiRAGeNews, MMFakeBench, and AMG datasets as the concept capacity increases. As illustrated in Figure~\ref{fig:conceptsense}, we observe that: (1) increasing the concept budget initially yields performance gains across all three datasets; (2) performance reaches an optimum when the maximum concept constraint is set to 5; and (3) performance degrades when the constraint exceeds 5.

\begin{figure}
    \centering
    \includegraphics[width=1\linewidth]{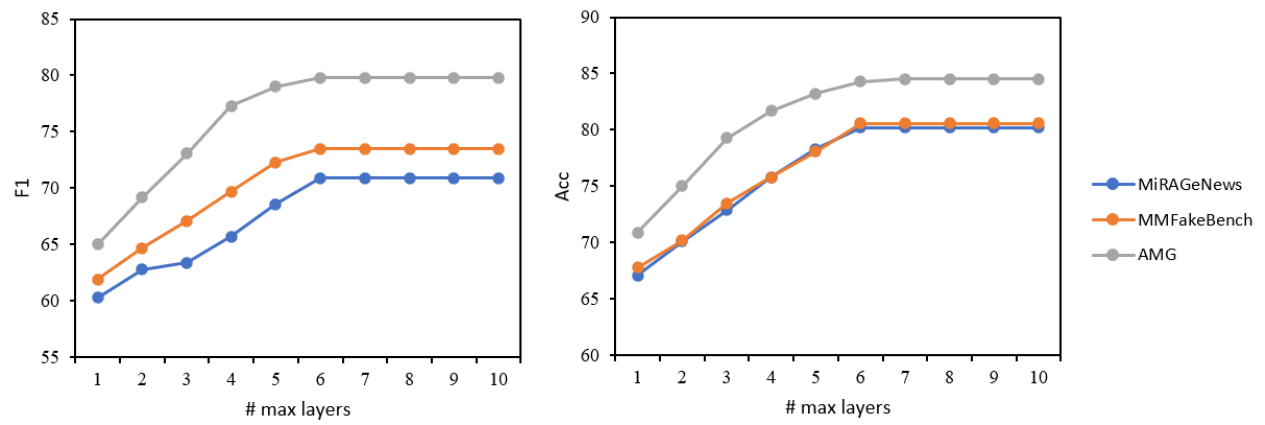}
    \caption{Impact of the max layers constraint (hierarchical depth).}
    \label{fig:layersense}
\end{figure}

To get an intuitive understanding of the hierarchical depth's impact, we evaluate PCGR's performance across varying numbers of concept layers. The F1 and Accuracy trajectories for the MiRAGeNews, MMFakeBench, and AMG datasets are illustrated in Figure~\ref{fig:layersense}. Our study reveals that: (1) PCGR's performance initially correlates positively with an increase in the maximum layer constraint; and (2) our model's performance saturates as the constraint reaches 6 layers, indicating that PCGR effectively captures necessary abstractions with moderate depth, eliminating the need for excessively deep concept graphs.

